\documentclass[runningheads]{llncs}
\usepackage{graphicx}

\usepackage{tikz}
\usepackage{comment}
\usepackage{amsmath,amssymb} 
\usepackage{color}
\usepackage{orcidlink}

\usepackage[accsupp]{axessibility}  

\usepackage{bbm}
\usepackage{multicol}
\usepackage{multirow}
\usepackage{booktabs}
\usepackage{enumerate}
\usepackage{paralist}
\usepackage{xcolor}
\usepackage{subcaption}
\usepackage{makecell}

\newcommand{\eg}{\emph{e.g.}}
\newcommand{\ie}{\emph{i.e.}}
\newcommand{\etal}{\textit{et al.}}

\begin{document}

\pagestyle{headings}
\mainmatter
\def\ECCVSubNumber{100}  

\title{Robust Object Detection With Inaccurate Bounding Boxes} 

\titlerunning{Robust Object Detection With Inaccurate Bounding Boxes}
%
\author{
Chengxin Liu\inst{1}\index{Liu,Chengxin} 
\and Kewei Wang\inst{1}\index{Wang,Kewei}
\and Hao Lu\inst{1}\index{Lu,Hao}
\and Zhiguo Cao\inst{1}\thanks{Corresponding author.}\index{Cao,Zhiguo}
\and Ziming Zhang\inst{2}\index{Zhang,Ziming}
}
\authorrunning{Liu et al.}
%
\institute{
School of Artificial Intelligence and Automation, \\Huazhong University of Science and Technology, China
\and
Worcester Polytechnic Institute, USA \\
\email{\{cx\_liu,zgcao\}@hust.edu.cn}}

\maketitle

\begin{abstract}
Learning accurate object detectors often requires large-scale training data with precise object bounding boxes. However, labeling such data is expensive and time-consuming. 
As the crowd-sourcing labeling process and the ambiguities of the objects may raise noisy bounding box annotations, the object detectors will suffer from the degenerated training data.
In this work, we aim to address the challenge of learning robust object detectors with inaccurate bounding boxes. Inspired by the fact that localization precision suffers significantly from inaccurate bounding boxes while classification accuracy is less affected, we propose leveraging classification as a guidance signal for refining localization results. Specifically, by treating an object as a bag of instances, we introduce an Object-Aware Multiple Instance Learning approach (OA-MIL), featured with object-aware instance selection and object-aware instance extension. 
The former aims to select accurate instances for training, instead of directly using inaccurate box annotations. The latter focuses on generating high-quality instances for selection. 
Extensive experiments on synthetic noisy datasets (\ie, noisy PASCAL VOC and MS-COCO) and a real noisy wheat head dataset demonstrate the effectiveness of our OA-MIL.
Code is available at \url{https://github.com/cxliu0/OA-MIL}.

\end{abstract}

\section{Introduction}

\begin{figure}[t]
\centering
\begin{minipage}[c]{0.56\textwidth}
    \centering
    \includegraphics[width=0.94\linewidth]{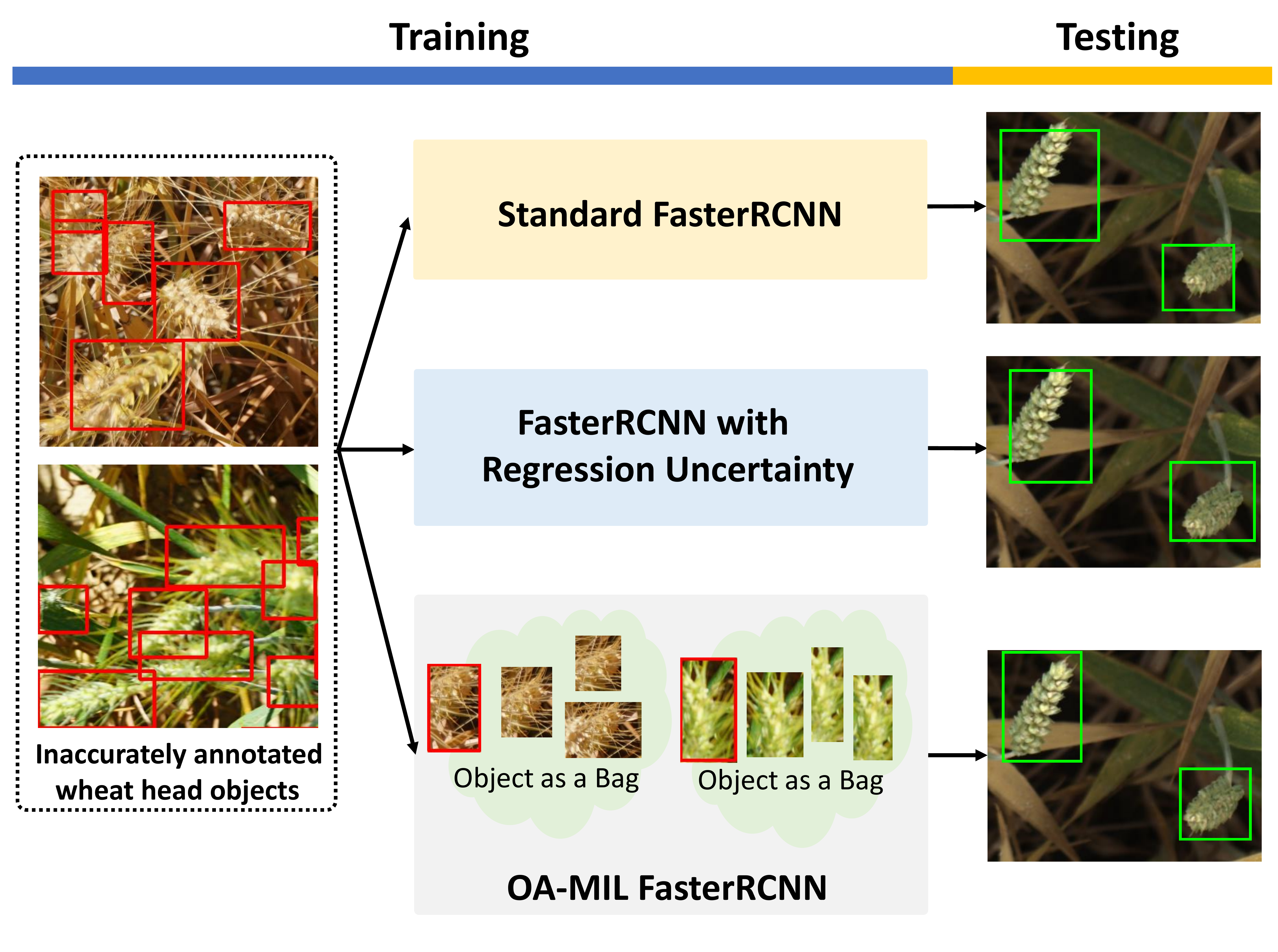}
	\caption{Illustration of standard FasterRCNN~\cite{ren2017fasterrcnn}, FasterRCNN with regression uncertainty~\cite{he2019kl}, and our OA-MIL FasterRCNN on the ECCV wheat head detection challenge dataset. Given inaccurately annotated objects, we aim to learn a robust object detector by treating each object as a bag of instances. The inaccurate ground-truth boxes are in red and the predictions are in green.}
	\label{fig:fig1_concept}
\end{minipage}
\quad
\begin{minipage}[c]{0.38\textwidth}
    \centering
    \includegraphics[width=1.0\linewidth]{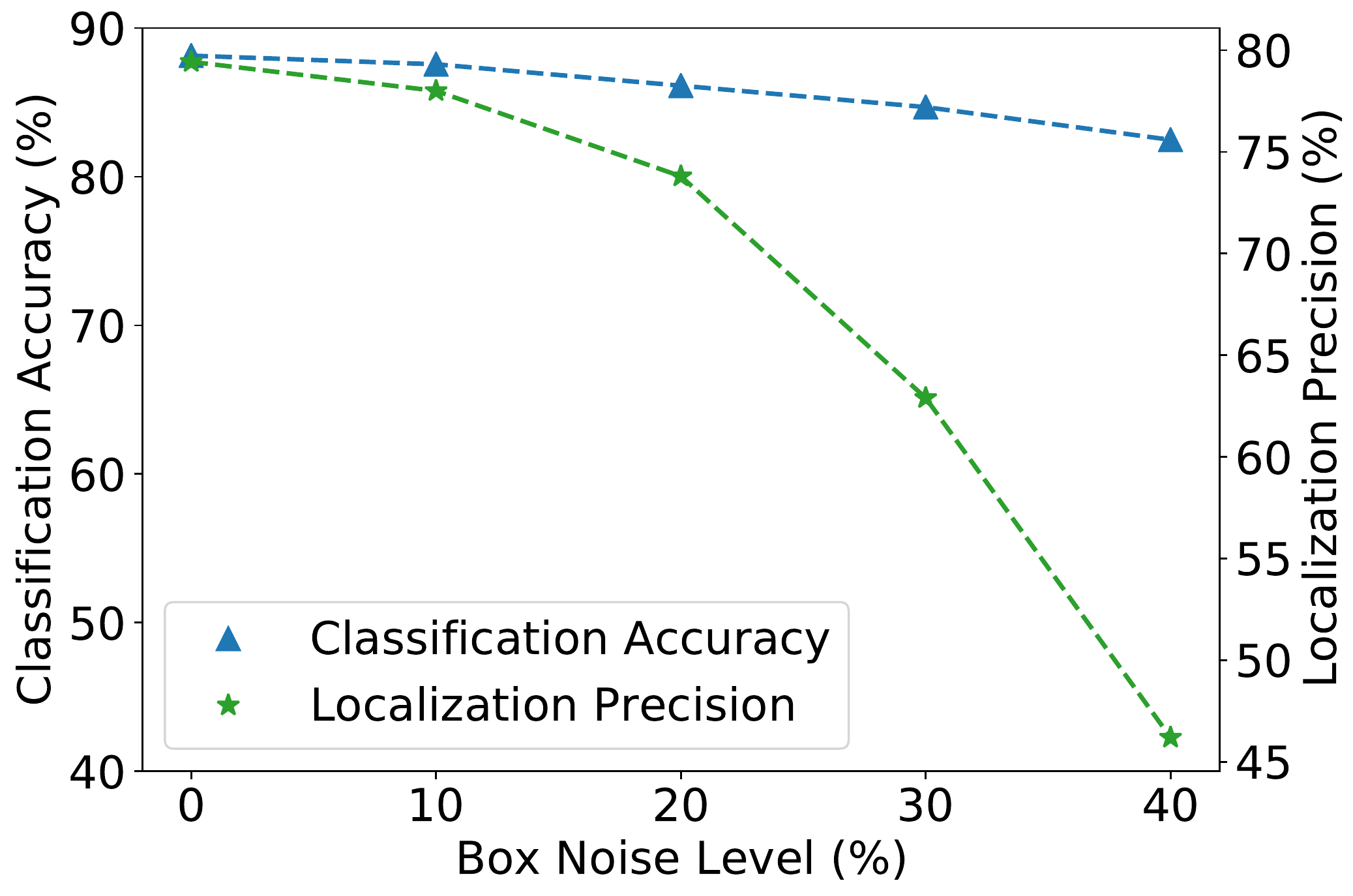}
	\caption{Classification accuracy and localization precision of FasterRCNN on the simulated ``noisy'' PASCAL VOC 2007 dataset~\cite{zisserman2010voc}, where box annotations are randomly perturbed. With the box noise level increases, \ie, ground-truth box becomes more and more inaccurate, localization precision drops significantly while the classification still maintains high accuracy.}
	\label{fig:cls_acc_plot}
\end{minipage}
\end{figure}

Despite remarkable progress has been witnessed in the field of object detection in recent years, the success of modern object detectors largely relies on large-scale datasets like ImageNet~\cite{deng2009imagenet} and MS-COCO~\cite{lin2014coco}. However, acquiring precise annotations is no easy task in professional and natural contexts.
In practical applications with professional backgrounds (\eg, agricultural crop observation and medical image processing), domain knowledge is often required to annotate objects. This situation leads to a dilemma, \ie, practitioners without computer vision background are not sure how to annotate high-quality boxes, while annotators without domain knowledge can also be difficult to annotate accurate object boxes. For example, recent wheat head detection challenge\footnote[1]{https://www.kaggle.com/c/global-wheat-detection\label{website}} that was hosted at the European Conference on Computer Vision (ECCV) workshop 2020 has shown that precise object bounding boxes are not easy to obtain, because in some domains the definition of the object is significantly different from generic objects in COCO, thus brings annotation ambiguities (Fig.~\ref{fig:fig1_concept}). In these cases, there will be a demand calling for algorithms dealing with noisy bounding boxes.
On the other hand, annotating a large amount of common objects in the natural context is expensive and time-consuming.
To reduce the annotation cost~\cite{xu2021mrnet}, 
dataset producers may rely on social media platforms or crowd-sourcing platforms.
Nevertheless, the above strategies would lead to low-quality annotations. Recent work~\cite{xu2021mrnet} argues that the object detectors will suffer from the degenerated data.
In addition, even large-scale datasets (\eg, MS-COCO) are dedicated annotated, box ambiguities~\cite{he2019kl} still exist. Therefore, tackling noisy bounding boxes is a practical and meaningful task.

Recently, learning object detectors with noisy data have gained a surge of interest, several approaches~\cite{bernhard2021,chadwick2019noisy,li2020noisy,xu2021mrnet} have made attempted to tackle noisy annotations. These approaches often assume that the noise occurs both on category labels and bounding box annotations, and devise a disentangled architecture to learn object detectors. 
Different from previous work, we focus on object detection with noisy bounding box annotations. The reasons are two-fold:
\begin{inparaenum}[i)]
    \item due to the ambiguities of the objects~\cite{he2019kl} and the crowd-sourcing labeling process, box noise commonly exists in the real world;
    \item object detection datasets~\cite{alina2020openimagev4} often involve object class verification, thus we consider noisy category labels are less severe than inaccurate bounding boxes.
\end{inparaenum} 

Motivated by the observation that localization precision suffers significantly from inaccurate bounding boxes while classification accuracy is less affected (Fig.~\ref{fig:cls_acc_plot}), we propose leveraging classification as a guidance signal for localization. Specifically, we present an Object-Aware Multiple Instance Learning approach by treating each object as a bag of instances, where the concept of the object bag is illustrated in Fig.~\ref{fig:fig1_concept}.
The idea is to select accurate instances from the object bags for training, instead of using inaccurate box annotations.
Our approach is featured with object-aware instance selection and object-aware instance extension. The former is designed to select accurate instances and the latter to generate high-quality instances for selection. The optimization process involves jointly training the instance selector, the instance classifier, and the instance generator.
To validate the effectiveness of our approach, we experiment on both synthetic noisy datasets (\ie, noisy PASCAL VOC 2007~\cite{zisserman2010voc} and MS-COCO~\cite{lin2014coco}) and real noisy wheat head dataset~\cite{david2020wheat,david2021wheat}.
The main contributions are as follows: 
\begin{itemize}
    \item We contribute a novel view for learning object detectors with inaccurate bounding boxes by treating an object as a bag of instances;
    \item We present an Object-Aware Multiple Instance Learning approach, featured by object-aware instance selection and object-aware instance extension;
    \item OA-MIL exhibits generality on off-the-shelf object detectors and obtains promising results on the synthetic and the real noisy datasets.
\end{itemize}

\section{Related Work}

\textbf{Learning with Noisy Labels.}
Training accurate DNNs under noisy labels has been an active research area. 
A major line of research focuses on the classification task, and develops various techniques to deal with noisy labels, such as sample selection~\cite{han2018coteach,jiang2018mentornet}, label correction~\cite{ma2018d2l,song2019selfie}, and robust loss functions~\cite{aritra2017mae,zhang2018gce}. 
Recently, much effort~\cite{bernhard2021,chadwick2019noisy,li2020noisy,mao2020icip,xu2021mrnet} has been devoted to the object detection task.
Simon \etal~\cite{chadwick2019noisy} first investigate the impact of different types of label noise on object detection, and propose a per-object co-teaching method to alleviate the effect of noisy labels.
On the other hand, Li \etal~\cite{li2020noisy} propose a learning framework that alternately performs noise correction and model training to tackle noisy annotations, where the noisy annotations consist of noisy category labels and noisy bounding boxes. 
Xu \etal~\cite{xu2021mrnet} further introduce a meta-learning based approach to tackle noisy labels by leveraging a few clean samples. 

In contrast to previous works, we emphasize learning object detectors with inaccurate bounding boxes and contribute a novel Object-Aware MIL view to addressing this problem. In addition, we do not assume the accessibility to clean box annotations as previous work~\cite{xu2021mrnet} does.

\noindent \textbf{Weakly-Supervised Object Detection (WSOD).}
WSOD refers to learning object detectors with only image-level labels. 
The majority of previous works formulate WSOD as a multiple instance learning (MIL) problem~\cite{dietterich1997mil}, where each image is considered as a ``bag'' of instances (instances are tentative object proposals) with image-level label. 
Under this formulation, the learning process alternates between detector training and object location estimation. Since MIL leads to a non-convex optimization problem, solvers may get stuck in local optima. Accordingly, much effort~\cite{bilen2015convex,cinbis2014mfold,deselaers2010app,siva2011drift,siva2012defence,song2014discovery} 
has been made to help the solution escape from local optima. 
Recently, deep MIL methods~\cite{bilen2016deepmil} emerge. However, the non-convexity problem remains. To address this problem, various techniques have been developed, including spatial regularization~\cite{bilen2016deepmil,diba2017casmil,wan2018minent}, context information~\cite{vadim2016contextloc,wei2018ts2c}, and optimization strategy~\cite{diba2017casmil,li2016domain,tang2017online,wan2018minent,wan2019cmil}.
For example, Zhang \etal~\cite{zhang2019sdloc} tackle noisy initialized object locations in WSOD and propose a self-directed localization network to identify noisy object instances.  

In this work, we tackle learning object detectors with noisy box annotations, which is different from WSOD where only image-level labels are given.
Despite we also formulate object detection as a MIL problem, 
we remark that our formulation is significantly different from WSOD in two aspects: i) we establish the concept of the bag on the object instead of on the image, which encodes object-level information; ii) we dynamically construct the object bag instead of using a fixed one as in WSOD, yielding a higher performance upper bound.

\medskip
\noindent \textbf{Semi-Supervised Object Detection (SSOD).}
SSOD aims to train object detectors with a large scale of image-level annotations and a few box-level annotations. 
Some prior works~\cite{tang2016semitransfer,tang2018semipami,uijlings2018semirevisit} address SSOD by knowledge transfer, where the information is transferred from source classes with bounding-box labels to target classes with only image-level labels. 
Other than the knowledge transfer paradigm, a recent work~\cite{gao2019notercnn} adopts a training-mining framework and proposes a noise-tolerant ensemble RCNN to eliminate the harm of noisy labels. 

However, previous SSOD methods generally assume the availability of clean bounding box annotations. In contrast, we only assume the accessibility to noisy bounding box annotations.


\section{Object-Aware Multiple Instance Learning}

In this work, we aim to learn a robust object detector with inaccurate bounding box annotations.
Motivated by the observation that classification maintains high accuracy under noisy box annotations (Fig.~\ref{fig:cls_acc_plot}), we suggest leveraging classification to guide localization.
Intuitively, instead of using the inaccurate ground-truth boxes, we expect the classification branch to select more precise boxes for training. This idea derives the concept of object bag, where each object is formulated as a bag of instances for selection.
Build upon the object bag, we present an Object-Aware Multiple Instance Learning approach that features object-aware instance selection and object-aware instance extension. 
In the following, we first introduce some preliminaries about MIL. Then, we present our Object-Aware Multiple Instance Learning formulation. Finally, we show how to deploy our method on modern object detectors like FasterRCNN~\cite{ren2017fasterrcnn} and RetinaNet~\cite{lin2017retina}.

\subsection{Preliminaries} \label{preliminary}

Given image-level labels, an MIL method~\cite{cinbis2014mfold,wan2019cmil} in WSOD treats each image as a bag of instances, where instances are tentative object proposals. The learning process alternates between instance selection and instance classifier learning.

Formally, let $B_i \in \mathbb{\mathcal{B}}$ denote the $i^{th}$ bag (image) and $\mathbb{\mathcal{B}}$ denote the bag set (all training images). Each $B_i$ is associated with a label $ y_i \in \{1,-1 \}$, where $y_i$ indicates whether $B_i$ contains positive instances.
Also, let $\mathbf{b}_i^{j}$ denote the $j^{th}$ instance of bag $B_i$ (\ie, $\mathbf{b}_i^{j}$ encodes the coordinates of an instance), where $j \in \{ 1,2, \ldots,N_i \}$ and $N_i$ is the number of instances in $B_i$. With the definitions above, the instance selector $f(\mathbf{b}_i^{j}, \omega_f)$ with parameter $\omega_f$ is applied to a positive bag $B_i$ to select the most positive instance $\mathbf{b}_i^{j^*}$, the index $j^*$ is obtained by:
\begin{gather}  
	\begin{aligned}
		j^* = \arg \mathop{\max}\limits_{j} f(\mathbf{b}_i^{j}, \omega_f), 
	\end{aligned}
	\label{eq:inst_selection}
\end{gather}
where the instance selector $f(\mathbf{b}_i^{j}, \omega_f)$ takes an instance $\mathbf{b}_i^{j}$ as input and outputs a confidence score that is in the range of $[-1,1]$.
Then, the selected instance $\mathbf{b}_i^{j^*}$ is used to train the instance classifier
$g(\mathbf{b}_i^{j}, \omega_g)$ with parameter $\omega_g$. The overall loss function is defined as:
\begin{gather}  
	\begin{aligned}
		L(\mathbb{\mathcal{B}}, \omega_f, \omega_g) = \sum_{i} L_{f}(B_i, \omega_f) + L_{g}(B_i, \mathbf{b}_i^{j^*}, \omega_g),
	\end{aligned}
	\label{eq:mil_train}
\end{gather}
where $L_{f}$ and $L_{g}$ are the loss of instance selector and instance classifier, respectively. Typically, $L_{f}$ is defined as a standard hinge loss:
\begin{gather}  
	\begin{aligned}
		L_{f}(B_i, \omega_f) = \max(0, 1 - y_i \mathop{\max}\limits_{j} f(\mathbf{b}_i^{j}, \omega_f)).
	\end{aligned}
	\label{eq:hinge_loss}
\end{gather}
And the instance classifier loss $L_g$ is defined as log-loss for classification. 

\begin{figure}[t]
	\centering
	\includegraphics[width=1.0\linewidth]{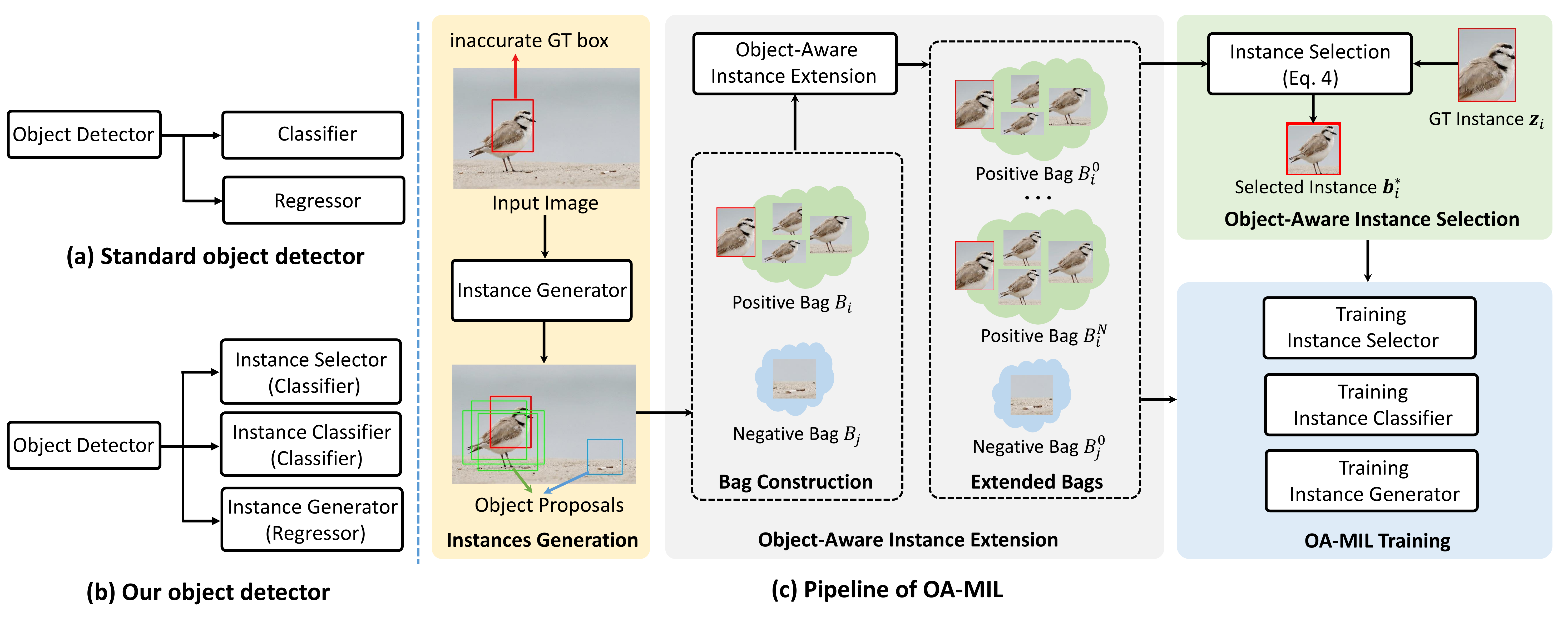}
	\caption{\textbf{An overview of our OA-MIL formulation}. We augment the standard object detector (a) with an instance selector, forming our object detector (b). (c) illustrates the pipeline of OA-MIL. We first construct object bags based on the outputs of the instance generator (green and blue boxes), where the inaccurate ground-truth box (red box) is formulated as a positive bag $B_i$ and the background box (blue box) is treated as a negative bag $B_j$. Then, object-aware instance extension is applied to obtain an extended bag set $\{ B_i^0, B_i^1, \ldots, B_i^N, B_j \}$. Based on this bag set and the noisy ground-truth instance $\mathbf{z}^i$, we adopt object-aware instance selection to select the best positive instance $\mathbf{b}_i^{*}$ for training object detector (including instance selector, instance classifier, and instance generator).}
	\label{fig:pipeline}
\end{figure}

\subsection{Object-Aware MIL Formulation} \label{sec:OAMIL}

Despite we formulate object detection as a MIL problem, we argue that the existing MIL paradigm in WSOD could not address the learning problem under noisy box annotations. First, since an image is defined as a bag in WSOD, the localization prior of objects is ignored. Second, the bags in WSOD are simply a collection of object proposals produced by off-the-shelf object proposal generators like selective search~\cite{uijlings2013ss}, which limits the detection performance.

Different from WSOD, in the context of our object bag, two challenges need to be solved:
\begin{inparaenum}[i)]
    \item \emph{how to select accurate instance in each object bag for training}; and
    \item \emph{how to generate high-quality instances for each object bag.}
\end{inparaenum}

To address the above challenges, we introduce an Object-Aware MIL formulation, which jointly optimizes the instance selector, the instance classifier, and the instance generator. 
In the following, we first give the definition of object bag. 
Then, we introduce object-aware instance selection and object-aware instance extension, where the former is designed to select the accurate instance, while the latter aims to produce a set of high-quality instances for selection.  
Finally, we describe how to train the instance selector, the instance classifier, and the instance generator. Fig.~\ref{fig:pipeline} illustrates the pipeline of our OA-MIL formulation. 

\smallskip
\noindent \textbf{Bag Definition.}
We reuse the bag symbol in Sec.~\ref{preliminary}. But the definition of bag is different, \ie, we treat each object as a bag. 
We denote $B_i \in \mathbb{\mathcal{B}}$ as the $i^{th}$ bag (object), and $\mathbb{\mathcal{B}}$ denotes the bag set (all objects in training images). A label $ y_i \in \{1,-1 \}$ is attached to each bag $B_i$.
As illustrated in Fig.~\ref{fig:pipeline}, we treat inaccurate ground-truth box as a positive bag $B_i$ with $y_i=1$, and the background box is treated as a negative bag $B_j$ with $y_j=-1$.
Suppose bag $B_i$ contains $N_i$ instances, we denote $\textbf{b}_i^{j}$ as the $j^{th}$ instance of bag $B_i$, where $j \in \{ 1, \ldots,N_i \}$. 
With object bag defined above, we naturally introduce Eq.~\eqref{eq:hinge_loss} to train the instance selector.

\smallskip
\noindent \textbf{Object-Aware Instance Selection.}
Since we treat each inaccurate ground-truth box as a bag of instances, the quality of the selected instance is essential for training an accurate object detector. Intuitively, we expect the selected instance covers the actual object as tight as possible. However, as the instance selector has poor discriminative ability in the early stage of training, the instance classifier and instance generator will inevitably suffer from the low-quality positive instance. In some cases, poor instance initialization could render failure during training. 
As the inaccurate ground-truth box provides a strong prior of object localization, we jointly consider it and the selected instance to obtain a more suitable positive instance for training.

Specifically, we denote $\mathbf{z}_{i}$ as the inaccurate ground-truth instance. We perform object-aware instance selection by merging $\mathbf{z}_{i}$ and $\mathbf{b}_i^{j^*}$ as follows:
\begin{equation}
	{\mathbf{b}_i^{*}} = \varphi (f(\mathbf{b}_i^{j^*}, \omega_f)) \cdot \mathbf{b}_i^{j^*} 
	+ (1-\varphi ( f(\mathbf{b}_i^{j^*}, \omega_f) ) ) \cdot \mathbf{z}_{i},
	\label{eq:oa_instance}
\end{equation}
where $\mathbf{b}_i^{j^*}$ is the most positive instance selected by the instance selector,
and $\varphi$ is a mapping function, which adaptively assigns the coefficient for $\mathbf{b}_i^{j^*}$ and $\mathbf{z}_{i}$.

Recall that our goal is to select high-quality positive instances for training, thus we expect $\varphi (\cdot)$ to satisfy two conditions.
First, higher weights should be assigned to $\mathbf{b}_i^{j^*}$ when $f(\mathbf{b}_i^{j^*}, \omega_f)$ has large value, because it indicates the confidence of the positive instance $\mathbf{b}_i^{j^*}$. Second, $\varphi (\cdot)$ should balance the weights of $\mathbf{b}_i^{j^*}$ and $\mathbf{z}^i$ instead of relying on $\mathbf{b}_i^{j^*}$ when $f(\mathbf{b}_i^{j^*}, \omega_f)$ is close to $1$. 
To satisfy the above conditions, we adopt a bounded exponential function as follows:
\setlength{\abovedisplayskip}{5pt}
\setlength{\belowdisplayskip}{4pt}
\begin{equation}
	\varphi (x) = \min (x^\gamma, \theta),
	\label{eq:exp_func}
\end{equation}
where $\gamma$ and $\theta$ are hyper-parameters, and $x \in [0, 1]$. 
A key property of Eq.~\eqref{eq:exp_func} is that the ascent speed and the upper bound of $\varphi$ are controllable. 

\medskip
\noindent \textbf{Object-Aware Instance Extension.}
The quality of instances is another factor that affects the training process.
In our formulation, 
the bags are dynamically constructed based on the outputs of the instance generator. Thus, the quality of bag instances can not always be guaranteed.
Fortunately, the instances inside a positive bag are homogeneous, \ie, instances are closely related to each other both on spatial location and class information. Therefore, it is possible to promote the quality of a positive bag by extending the positive instances.

We present two strategies for instance extension.
The first strategy is to obtain new positive instances by recursively constructing positive bags. Specifically, we first obtain the initial object bag based on the noisy ground-truth boxes, then we use the most positive instance selected by Eq.~\eqref{eq:oa_instance} to construct a new positive bag. The process repeats until reaching the termination condition. This strategy is generic and applicable to any existing object detectors. The second strategy is to refine the positive instances in a multi-stage manner~\cite{cascadercnn}, which is suitable for object detectors that feature with a bounding box refinement module (\eg, FasterRCNN~\cite{ren2017fasterrcnn}). 
The extended object bags are subsequently used to train the instance selector. Note that we do not extend negative bags.

Suppose we have conducted $N$ times of instance extension, which produces a set of extended positive bags $\{ B_i^0, B_i^1, \ldots, B_i^N \}$, where $B_i^0$ denotes the initial object bag $B_i$ (As shown in Fig.~\ref{fig:pipeline}). We utilize the extended object bags to optimize the instance selector, the loss thus becomes:
\begin{equation}
    L_{f}(\{B_i^k\}, \omega_f) = \sum_{k} L_{f}(B_i^k, \omega_f), \\
    \label{eq:inst_extention}
\end{equation}
where $k \in \{0, 1, \ldots, N \}$ only if $B_i$ is a positive bag.

\medskip
\noindent \textbf{OA-MIL Training.}
Our OA-MIL involves jointly optimizing the instance selector, instance classifier, and the instance generator. The instance selector is trained using Eq.~\eqref{eq:inst_extention}.
As the instance classifier $g$ (with parameter $\omega_g$) is used to classify object, we adopt the binary-log-loss to train it:
\begin{gather}  
	\begin{aligned}
		L_{g}(B_i, \mathbf{b}_i^{*}, \omega_g) = - \sum_{j} \log(y_{i,j} \cdot ( g(\mathbf{b}_i^{j}, \omega_g ) - \frac{1}{2}) + \frac{1}{2} ), 
	\end{aligned}
	\label{eq:ce_loss}
\end{gather}
where $g(\mathbf{b}_i^{j}, \omega_g ) \in (0,1)$, which represents the probability of $\mathbf{b}_i^{j}$ contains objects with positive class. $y_{i,j}$ is defined as:
\begin{equation}
	\begin{split}
		y_{i,j} = 
		\begin{cases}
			+1, \, \text{if} \, y_i=1 \, \text{and} \, \text{IoU}(\mathbf{b}_i^{j}, \mathbf{b}_i^{*}) \geq 0.5  \\
			-1, \, \text{if} \, y_i=1 \, \text{and} \, \text{IoU}(\mathbf{b}_i^{j}, \mathbf{b}_i^{*}) < 0.5 \\
			-1, \, \text{if} \, y_i=-1 \\
		\end{cases}\, ,
	\end{split}
	\label{eq:inst_label}
\end{equation}
where $\text{IoU}$ denotes the Intersection over Union between two instances. Note that the loss of each instance becomes $\log(g(\mathbf{b}_i^{j}, \omega_g))$ when $y_{i,j}=1$, and $\log(1 - g(\mathbf{b}_i^{j}, \omega_g))$ otherwise.

One major difference between our MIL and WSOD MIL is that we jointly train a learnable instance generator, which is crucial for dealing with inaccurate bounding boxes.
The loss function of the instance generator is as follows: 
\begin{gather}  
	\begin{aligned}
		L_{\phi} (\mathbb{\mathcal{B}}, \omega_\phi) = \sum_{i} \mathbbm{1}(B_i) \cdot L_{reg} (B_i, \mathbf{b}_i^{*}, \omega_\phi), \\
	\end{aligned}
	\label{eq:det_loss}
\end{gather}
where $\omega_\phi$ is the parameters of instance generator, $\mathbbm{1}(B_i)$ equals $1$ if $B_i$ is a positive bag, otherwise $\mathbbm{1}(B_i)$ is $0$ because a negative bag does not correspond to any actual objects, and $L_{reg}$ is defined as:
\begin{equation}
	L_{reg} (B_i, \mathbf{b}_i^{*}, \omega_\phi) = \sum_{j} \delta (y_{i,j}) \cdot \ell_{reg} (\mathbf{b}_i^{j}, \mathbf{b}_i^{*}),
	\label{eq:reg_loss}
\end{equation}
where $\delta(\cdot)$ is the unit-impulse function ($\delta(y_{i,j})$ equals to $1$ when $y_{i,j}$ is $1$, otherwise $0$), and $\ell_{reg}$ is a regression loss like $\ell_1$ loss or smooth $\ell_1$ loss~\cite{ren2017fasterrcnn}.

To summarize, the overall loss function is formulated as:
\begin{gather}  
	\begin{aligned}
		L (\mathbb{\mathcal{B}}, \omega_f, \omega_g, \omega_\phi) = \lambda \sum_{i} \sum_{k} L_{f}(B_i^k, \omega_f) + \sum_{i} L_{g}(B_i^{0}, \mathbf{b}_i^{*}, \omega_g) \\ 
		+ \sum_{i} \mathbbm{1}(B_i^0) \cdot L_{reg} (B_i^0, \mathbf{b}_i^{*}, \omega_\phi), \\
	\end{aligned}
	\label{eq:oa_mil_loss}
\end{gather}
where $\lambda$ is a balance parameter, $\mathbb{\mathcal{B}}$ is the extended object bags set, $B_i^0$ is the initial object bag, and $\mathbf{b}_i^{*}$ is selected by Eq.~\eqref{eq:oa_instance}.

\subsection{Deployment to Off-the-Shelf Object Detectors}

We remark that our formulation is general and is not limited to specific object detectors. To demonstrate the generality of our method, we apply our method on a two-stage detector---FasterRCNN~\cite{ren2017fasterrcnn} and a one-stage detector---RetinaNet~\cite{lin2017retina}. 
Here we introduce the deployment procedure on FasterRCNN. 
It includes two steps, the first step is to construct object bags and the second step is to apply OA-MIL on FasterRCNN. 
More details can be found in the supplementary.

\smallskip
\noindent \textbf{Bag Construction.}
We construct object bags based on the outputs of the second stage of FasterRCNN. Specifically, we treat each inaccurately annotated object as a positive bag, where instances are positive anchors (object proposals) corresponding to a specific object.

\smallskip
\noindent \textbf{OA-MIL Deployment.}
The instance selector and the instance classifier share the same classifier. The regressor in the second stage is treated as the instance generator. 
We perform object-aware instance extension by multi-stage refinement, which produces a set of extended object bags $\{B_i^0, B_i^1, \ldots, B_i^{N}\}$. Then, object-aware instance selection is applied to select the best instance $\mathbf{b}_{i}^{*}$, where $\mathbf{b}_i^{*}$ is used to train the instance generator (regressor), the instance selector (classifier), and the object detector (classifier). We follow Eq.~\eqref{eq:oa_mil_loss} to train the second stage of FasterRCNN. Note that the training objective of RPN is the same as~\cite{ren2017fasterrcnn}.

\smallskip
\noindent \textbf{Implementation Details.}
We implement our method on FasterRCNN~\cite{ren2017fasterrcnn} with ResNet50-FPN~\cite{he2016resnet,lin2017fpn} backbone. 
Following common practices~\cite{ross2018detectron}, the model is trained with ``$1\times$'' schedule.
The hyper parameters are set as $\gamma=7.5$, $\theta=0.85$, $N=4$, and $\lambda$ is selected from $\{0.01, 0.1\}$ (depending on datasets and noise levels).
Our implementation is based on MMDetection toolbox~\cite{chen2019mmdet}.

\section{Results and Discussions}

\subsection{Datasets and Evaluation Metrics} \label{sec:dataset}

\begin{figure}[t]
    \centering
  \begin{subfigure}{0.38\textwidth}
    \includegraphics[width=1.0\linewidth]{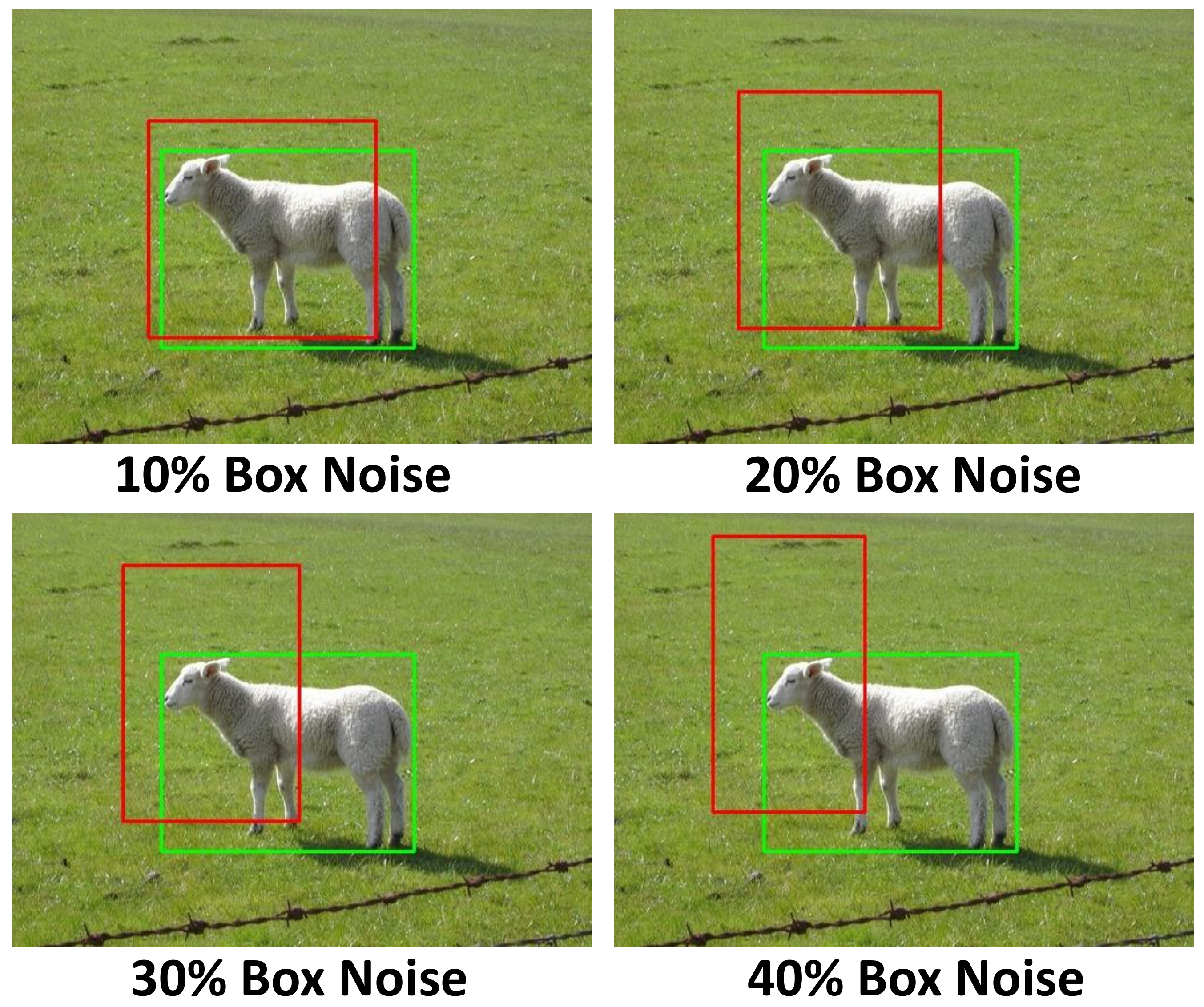}
	\caption{Noisy VOC dataset}
	\label{fig:voc_noise_boxes}
  \end{subfigure}
  \quad
  \begin{subfigure}{0.38\textwidth}
    \centering
    \includegraphics[width=0.87\linewidth]{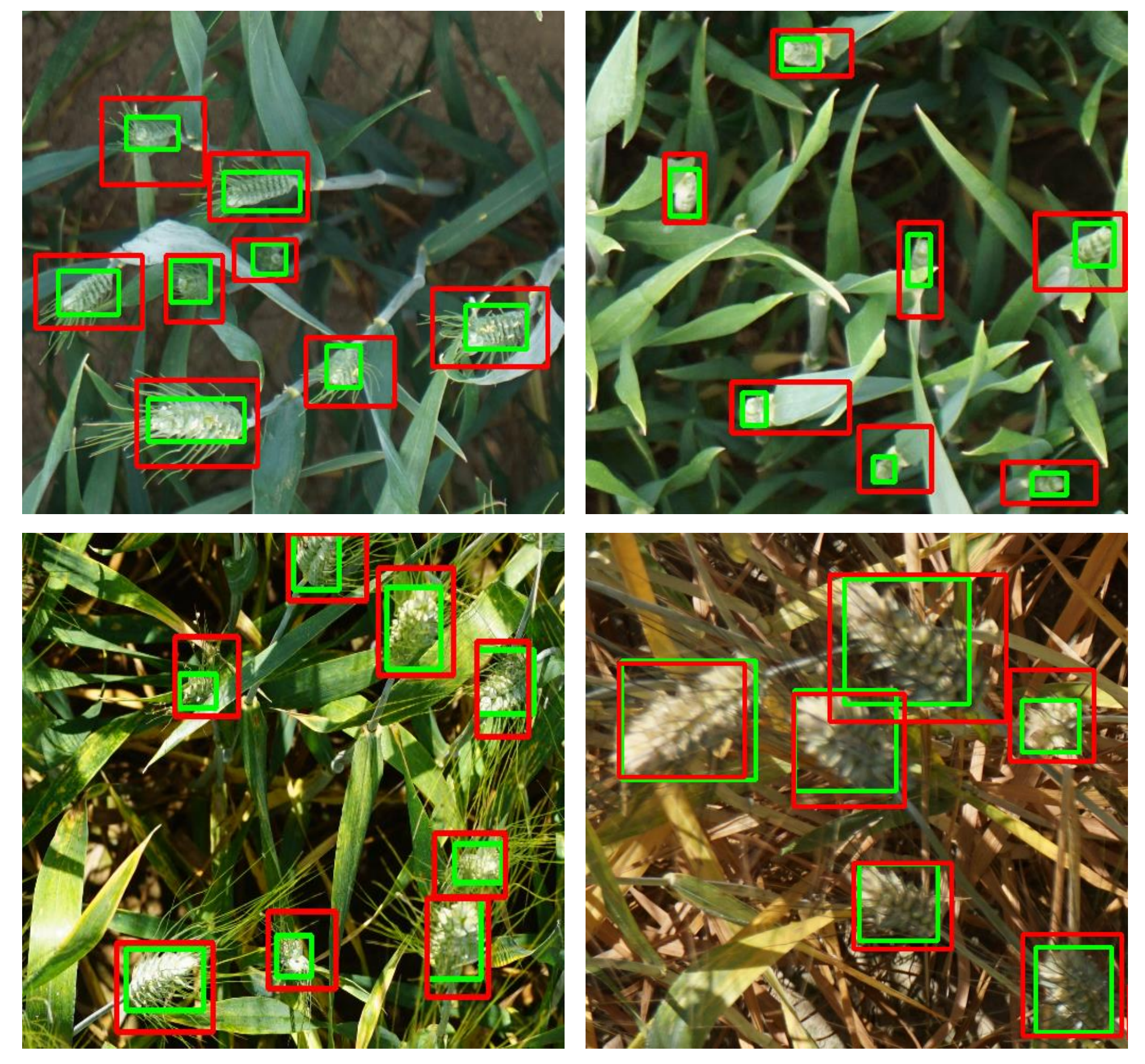}
    \caption{GWHD dataset} 
    \label{fig:wheat_noise_boxes}
  \end{subfigure}
\caption{\textbf{Examples of the inaccurate bounding boxes (red boxes) on VOC and GWHD dataset}. The clean ground-truth boxes are in green.}
\end{figure}

\subsubsection{Synthetic Noisy Dataset.} 
Modern object detection datasets are delicately annotated and contain few inaccurate bounding boxes. Thus, we simulate noisy bounding boxes by perturbing the clean ones on two object detection datasets, including PASCAL VOC 2007~\cite{zisserman2010voc} and MS-COCO~\cite{lin2014coco}. 

\textit{Box Noise Simulation.} We simulate noisy bounding boxes by perturbing clean boxes. Specifically, let $(cx,cy,w,h)$ denote the center $x$ coordinate, center $y$ coordinate, width, and height of an object. We simulate an inaccurate bounding box by randomly shifting and scaling the box as follows:
\begin{equation}
	\begin{split}
		\begin{cases}
			\hat{cx} = cx + \Delta_x \cdot w, \quad \hat{cy} = cy + \Delta_y \cdot h, \\
			\hat{w} = (1 + \Delta_w) \cdot w, \quad \hat{h} = (1 + \Delta_h) \cdot h,\\
		\end{cases}
	\end{split}
	\label{eq:box_noise}
\end{equation}
where $\Delta_x$, $\Delta_y$, $\Delta_w$, and $\Delta_h$ follow the uniform distribution $U(-r,r)$, $r$ is the box noise level. We simulate various box noise levels ranging from $10\%$ to $40\%$. 
For example, when $r=40\%$, $\Delta_x$, $\Delta_y$, $\Delta_w$, and $\Delta_h$ are in the range of $(-0.4,0.4)$.
Note that Eq.~\eqref{eq:box_noise} is performed on \textbf{every bounding box} in the training data. 
Fig.~\ref{fig:voc_noise_boxes} shows examples of the synthetic inaccurate bounding boxes under different box noise level $r$'s on the VOC dataset, where $r$ ranges from $10\%$ to $40\%$. 

\medskip
\noindent \textbf{Real Noisy Dataset.} 
We also evaluate our approach on the Global Wheat Head Detection (GWHD) dataset~\cite{david2020wheat,david2021wheat}.
This dataset includes $3.6\text{K}$ training images, $1.4\text{K}$ validation images, and $1.3\text{K}$ test images.
It has two versions of training data, the first ``noisy'' challenge version\textsuperscript{\ref{website}} (inaccurate bounding box annotations) and  the second ``clean'' version (calibrated clean annotations).
Specifically, the ``noisy'' version contains around $20\%$ noisy ground-truth boxes and the rest $80\%$ boxes are the same as the ``clean'' version.
We separately train on the ``noisy'' and ``clean'' training data to validate our approach. Fig.~\ref{fig:wheat_noise_boxes} shows some examples of the real inaccurate bounding boxes and the calibrated clean boxes. 

\medskip
\noindent \textbf{Evaluation Metric.} 
For VOC and COCO, we use mean average precision (mAP@.5) and mAP@[.5,.95] as evaluation metrics. Regarding GWHD dataset, we follow the GWHD Challenge 2021\footnote[2]{https://www.aicrowd.com/challenges/global-wheat-challenge-2021\label{gwc2021}} to use Average Domain Accuracy (ADA) as the evaluation metric. 

\begin{table}[!t]
	\caption{Performance comparison on the PASCAL VOC 2007 test set. The evaluation metric is mAP@0.5 (\%). The best performance is in \textbf{boldface}.}
	\label{table:voc_sota}
	\centering
	\setlength{\tabcolsep}{2.5mm}
	\begin{tabular}{l| l | c c c c}
		\toprule
		\multirow{2}{*}{Model}  & \multirow{2}{*}{Method} 
		& \multicolumn{4}{c}{Box Noise Level} \\
		& & 10\% & 20\% & 30\% & 40\%\\
		
		\midrule
		\multirow{5}{*}{FasterRCNN} 
		& Noisy-FasterRCNN & 76.3 & 71.2 & 60.1 & 42.5 \\
		& KL loss~\cite{he2019kl} & 75.8 & 72.7 & 64.6 & 48.6\\
		& Co-teaching~\cite{han2018coteach} & 75.4 & 70.6 & 60.9 & 43.7 \\
		& SD-LocNet~\cite{zhang2019sdloc} & 75.7 & 71.5 & 60.8 & 43.9 \\
		
		\cmidrule{2-6}
		& Ours & \textbf{77.4} & \textbf{74.3} & \textbf{70.6} & \textbf{63.8} \\

		\midrule
		\multirow{3}{*}{RetinaNet} 
		& Noisy-RetinaNet & 71.5 & 67.5 & 57.9 & 45.0 \\
		& FreeAnchor~\cite{zhang2019free} & 73.0 & 67.5 & 56.2 & 41.6 \\
		\cmidrule{2-6}
		& Ours & \textbf{73.1} & \textbf{69.1} & \textbf{62.9} & \textbf{53.4} \\

        \midrule
        \midrule
		\multirow{2}{*}{Clean Model} 
		& Clean-FasterRCNN & 77.2 & 77.2 & 77.2 & 77.2 \\
		& Clean-RetinaNet & 73.5 & 73.5 & 73.5 & 73.5 \\
		
		\bottomrule
		
	\end{tabular}
\end{table}

\subsection{Comparison With State of the Art}

We compare our method with several state-of-the-art approaches~\cite{han2018coteach,he2019kl,zhang2019sdloc} on PASCAL VOC 2007~\cite{zisserman2010voc}, MS-COCO~\cite{lin2014coco}, and GWHD~\cite{david2020wheat,david2021wheat} datasets. 
Note that, we denote Clean-FasterRCNN and Noisy-FasterRCNN as FasterRCNN models trained under clean and noisy training data with the default setting, respectively. Similarly, Clean-RetinaNet and Noisy-RetinaNet denote RetinaNet models trained under clean and noisy training data, respectively.

\medskip
\noindent \textbf{Results on the VOC 2007 dataset.} 
Table~\ref{table:voc_sota} shows the comparison results on the VOC 2007 test set.
For FasterRCNN, we observe that inaccurate bounding box annotations significantly deteriorate the detection performance of the vanilla model. On the contrary, our approach is more robust to noisy bounding boxes and outperforms other methods by a large margin under high box noise levels, \eg, $30\%$ and $40\%$ box noise. 
In addition, Co-teaching and SD-LocNet only slightly improve the detection performance, which indicates that small-loss sample selection and sample weight assignment can not well tackle noisy box annotations. 
For RetinaNet, we compare our approach with the vanilla RetinaNet model and FreeAnchor~\cite{zhang2019free}. As shown in Table~\ref{table:voc_sota}, our approach still achieves consistent improvement over the vanilla model, which indicates that our approach is effective on both two-stage and one-stage detectors.

\begin{table}[t]
	\centering
	\setlength{\tabcolsep}{0.1mm}
	\caption{Performance comparison on the MS-COCO dataset.}
	\label{table:coco_sota}
	\begin{tabular}{l| c c c c c c| c c c c c c}
		\toprule			
		\multirow{2}{*}{Method} & \multicolumn{6}{c|}{20\% Box Noise Level} 
		& \multicolumn{6}{c}{40\% Box Noise Level} \\
		
		& AP & $AP^{50}$  & $AP^{75}$ & $AP^{S}$ & $AP^{M}$ & $AP^{L}$ & AP & $AP^{50}$  & $AP^{75}$ & $AP^{S}$ & $AP^{M}$ & $AP^{L}$ \\
		
		\midrule
		\midrule
		\multicolumn{13}{c}{\textbf{FasterRCNN}} \\
		\midrule
		Clean-FasterRCNN & 37.9 & 58.1 & 40.9 & 21.6 & 41.6 & 48.7 & 37.9 & 58.1 & 40.9 & 21.6 & 41.6 & 48.7 \\
		Noisy-FasterRCNN & 30.4 & 54.3 & 31.4 & 17.4 & 33.9 & 38.7 & 10.3 & 28.9 & 3.3 & 5.7 & 11.8 & 15.1 \\
		KL loss~\cite{he2019kl}  & 31.0  & 54.3 & 32.4 & 18.0 & 34.9 & 39.5 & 12.1 & 36.7 & 3.7 & 6.2 & 13.0 & 17.4 \\
		Co-teaching~\cite{han2018coteach}  & 30.5 & 54.9 &  30.5 & 17.3 & 34.0 & 39.1 & 11.5 & 31.4 & 4.2 & 6.4 & 13.1 & 16.4 \\
		SD-LocNet~\cite{zhang2019sdloc}  & 30.0 & 54.5 & 30.3 & 17.5 & 33.6 & 38.7  & 11.3 & 30.3 & 4.3 & 6.0 & 12.7 & 16.6\\
		\midrule
		Ours & \textbf{32.1} & \textbf{55.3} & \textbf{33.2} & \textbf{18.1} & \textbf{35.8} & \textbf{41.6} & \textbf{18.6} & \textbf{42.6} & \textbf{12.9} & \textbf{9.2} & \textbf{19.9} & \textbf{26.5} \\

		\midrule
		\midrule
		\multicolumn{13}{c}{\textbf{RetinaNet}} \\
		\midrule
		Clean-RetinaNet & 36.7 & 56.1 & 39.0 & 21.6 & 40.4 & 47.4 & 36.7 & 56.1 & 39.0 & 21.6 & 40.4 & 47.4 \\
		Noisy-RetinaNet & 30.0 & 53.1 & 30.8 & 17.9 & 33.7 & 38.2 & 13.3 & 33.6 & 5.7 & 8.4 & 15.9 & 18.0 \\
		FreeAnchor~\cite{zhang2019free} & 28.6 & 53.1 & 28.5 & 16.6 & 32.2 & 37.0 & 10.4 & 28.9 & 3.3 & 5.8 & 12.1 & 14.9 \\
		\midrule
		Ours & \textbf{30.9} & \textbf{54.0} & \textbf{32.3} & \textbf{18.5} & \textbf{34.9} & \textbf{39.6} & \textbf{19.2} & \textbf{45.2} & \textbf{12.0} & \textbf{11.3} & \textbf{23.0} & \textbf{24.9} \\
		  
		\bottomrule
	\end{tabular}
\end{table}

\medskip
\noindent \textbf{Results on the MS-COCO Dataset.}
The comparison results on the MS-COCO dataset are reported in Table~\ref{table:coco_sota}. 
For FasterRCNN, our approach achieves considerable improvements over the vanilla model and performs favorably against state-of-the-art methods. 
For example, under $40\%$ box noise, the vanilla model suffers from catastrophic performance drop, \eg, $AP^{50}$ drops from $58.1$ to $28.9$. On the other hand, our approach significantly boosts the detection performance across all metrics, achieving $8.3\%$, $13.7\%$, and $9.6\%$ improvements on $AP$, $AP^{50}$, and $AP^{75}$, respectively. Co-teaching and SD-LocNet, however, still can not well address inaccurate bounding box annotations, but KL Loss slightly improves the performance under $20\%$ and $40\%$ box noise.
In addition, we observe that objects with different sizes suffer similarly under different noise levels.
For RetinaNet, our approach also obtains consistent improvements. For example, our approach improves the performance of the vanilla RetinaNet by $5.9\%$, $11.6\%$, and $6.3\%$ on $AP$, $AP^{50}$, and $AP^{75}$ under $40\%$ box noise, respectively.

\begin{table}[t]
	\caption{Comparison results on the GWHD validation and test set, where models are trained on ``noisy'' and ``clean'' training data, respectively. The evaluation metric is ADA. FRCNN denotes FasterRCNN.}	
	\label{table:wheat_sota}
	\centering
	\setlength{\tabcolsep}{0.1mm}
	\begin{tabular}{l | l| c c| c c}
		\toprule
		\multirow{2}{*}{Model} & \multirow{2}{*}{Method} 
		& \multicolumn{2}{c|}{Trained on ``Noisy'' GWHD}
		& \multicolumn{2}{c}{Trained on ``Clean'' GWHD} \\
		& & Val ADA & Test ADA & Val ADA  & Test ADA \\
		
		\midrule
		\multirow{5}{*}{FRCNN} 
		& Vanilla FRCNN & 0.608 & 0.509 & 0.632 & 0.511 \\
		& KL Loss~\cite{he2019kl} & 0.607 & 0.496 & 0.631 & 0.507  \\
		& Co-teaching~\cite{han2018coteach} & 0.624 & 0.491 & 0.631 & 0.504 \\
		& SD-LocNet~\cite{zhang2019sdloc} & 0.621 & 0.498 & 0.626 & 0.512\\
		\cmidrule{2-6}
		& Ours & \textbf{0.639} & \textbf{0.526} & \textbf{0.658} & \textbf{0.530}\\
		
		\midrule
		\multirow{3}{*}{RetinaNet} 
		& Vanilla RetinaNet & 0.607 & 0.494 & 0.622 & 0.503 \\
		& FreeAnchor~\cite{zhang2019free} & 0.619 & \textbf{0.517} & 0.635 & 0.525 \\
		\cmidrule{2-6}
		& Ours & \textbf{0.621} & 0.516 & \textbf{0.640} & \textbf{0.527}\\
		\bottomrule
	\end{tabular}
\end{table}

\medskip
\noindent \textbf{Results on the GWHD Dataset.}
Here we report the results on both noisy and clean training data. 

\textit{Results on the ``Noisy'' GWHD dataset.}
Table~\ref{table:wheat_sota} shows the comparison results.
Deploying our method on FasterRCNN boosts the Val ADA of the vanilla model from $0.608$ to $0.639$ and Test ADA from $0.509$ to $0.526$. Interestingly, our approach even performs better than the vanilla model that trained on clean training data ($0.639$ vs. $0.632$ on Val ADA, $0.526$ vs. $0.511$ on Test ADA). In addition, Co-teaching and SD-LocNet improve the Val ADA but deteriorate the Test ADA, we infer that the large domain gap between validation and test data leads to the controversy. For RetinaNet, our approach obtains moderate improvements and performs favorably against FreeAnchor. Note that FreeAnchor performs well on ``Noisy'' GWHD because the clean ground-truth boxes dominate this dataset (around $80\%$ boxes are clean), which is different from the synthetic VOC and COCO datasets where noisy ground-truth boxes are the majority.

\textit{Results on the ``Clean'' GWHD dataset.} 
Table~\ref{table:wheat_sota} shows that our approach can further improve the detection performance when trained on clean data.
The reason may be that our OA-MIL exploits the information between object instances, thus strengthening the discriminative capability of the detection model.
Specifically, we advance the performance of the vanilla FasterRCNN by $2.6\%$ and $1.7\%$ on Val ADA and Test ADA, respectively. 
Regarding RetinaNet, our approach outperforms the vanilla model by $1.8\%$ on Val ADA and $2.4\%$ on Test ADA, respectively. 
In addition, our approach can also cooperate with the runner-up solution~\cite{liu2021iccvw,liu2022pp} of the GWHD2021 challenge\textsuperscript{\ref{gwc2021}}, which adopts the idea of dynamic network~\cite{lu2022index} to improve wheat head detection.

\begin{table}[t]
	\centering
	\setlength{\tabcolsep}{0.8mm}
	\caption{Ablation study on the VOC 2007 test set and COCO validation set. The evaluation metric is mAP@0.5 (\%).}
	\label{table:ablation}
	\begin{tabular}{l| l| c c c| c c c c| c c}
		\toprule			
		\multirow{2}{*}{No.} & \multirow{2}{*}{Method} & \multirow{2}{*}{IS Loss} & \multirow{2}{*}{OA-IS} & \multirow{2}{*}{OA-IE} & \multicolumn{4}{c|}{VOC 2007} & \multicolumn{2}{c}{COCO} \\
		& & & & & 10\%  & 20\%  & 30\%  & 40\%  & 20\%  & 40\% \\
		
		\midrule
		B1 & Vanilla FasterRCNN & & & & 76.3 & 71.2 & 60.1 & 42.5 & 54.3 & 28.9 \\
		
		\midrule
		B2 & \multirow{3}{*}{OA-MIL FasterRCNN} 
		& \checkmark & & & 77.1 & 73.3 & 66.9 & 56.0 & 54.6 & 32.6\\
		B3 & & \checkmark & \checkmark & & 77.2 & 74.2 & 70.2 & 63.3 & 55.2 & 39.8\\
		B4 & & \checkmark & \checkmark & \checkmark & 77.4 & 74.3 & 70.6 & 63.8 & 55.3 & 42.6\\
		\bottomrule
		
	\end{tabular}
\end{table}

\subsection{Ablation Study}
Here we investigate the effectiveness of each component in our approach, including:
\begin{inparaenum}[(i)]
\item our object bag formulation, \ie, training object detector with instance selection loss (IS Loss), where the loss is computed based on object bag;
\item object-aware instance selection (OA-IS);
\item object-aware instance extension (OA-IE).
\end{inparaenum}
Table~\ref{table:ablation} shows the results. 
B1 is the performance of the vanilla FasterRCNN trained under different box noise levels. From B2 to B4, we gradually add IS Loss, OA-IS, and OA-IE into training. 
In addition, the analysis of parameter sensitivity (\eg, $\gamma$ and $\theta$ in Eq.~\eqref{eq:exp_func}) can be found in the supplementary.

\smallskip
\noindent \textbf{Effectiveness of Objec Bag Formulation.} 
Interestingly, simply training under our object bag formulation significantly boosts the mAP performance of FasterRCNN on the VOC 2007 dataset across several box noise levels. For instance, our object bag formulation achieves $6.8\%$ and $13.5\%$ improvements under $30\%$ and $40\%$ box noise level, respectively. As for the COCO dataset, we still obtains moderate improvements. 
An intuitive explanation is that the instance selector is forced to select high-quality instances (\eg, instance that covers the actual object more tightly) to minimize the loss function. As a consequence, the object detector benefits from the joint optimization process.

\begin{figure}[t]
	\centering
	\includegraphics[width=0.8\linewidth]{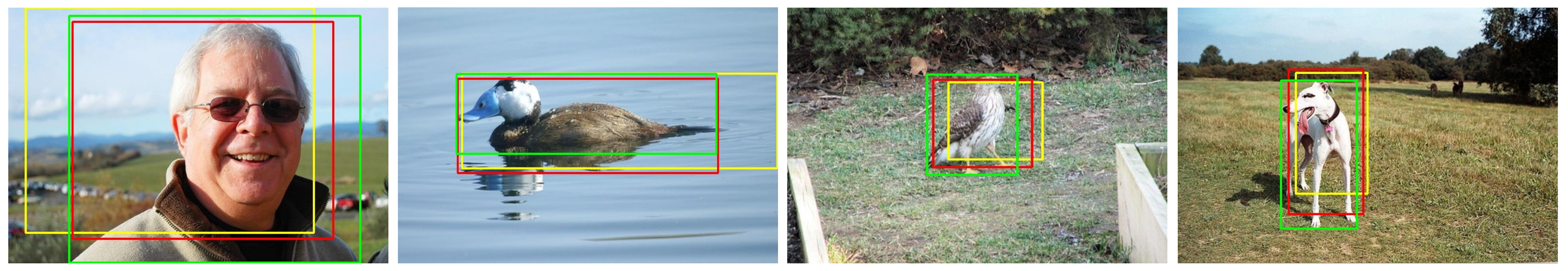}
	\caption{\textbf{Examples of the selected instances (red boxes)}. Noisy ground-truth boxes are in yellow and the clean ground-truth boxes are in green.}
	\label{fig:BLI_vis}
\end{figure}

\smallskip
\noindent \textbf{Effectiveness of OA-IS.} 
Applying OA-IS further improves the detection performance on the VOC and COCO datasets, especially under high box noise levels. For example, under 40\% box noise level, OA-IS boosts the performance from $56.0$ to $63.3$ on the VOC dataset and from $32.6$ to $39.8$ on the COCO dataset. To understand OA-IS more intuitively, we visualize the instances selected by OA-IS in Fig.~\ref{fig:BLI_vis}. 
It is clear that the selected instances cover the objects more tightly than the noisy ground-truth boxes. Although the selected instances are not perfect, they provide more precise supervision signals for training the instance classifier and the instance generator.

\medskip
\noindent \textbf{Effectiveness of OA-IE.} 
OA-IE is designed to improve the quality of the bag instances. We observe that the impact of OA-IE is minor under low box noise levels. The reason is likely that the quality of bag instances is relatively high under low box noise situations. Nevertheless, OA-IE still brings improvement under high noise levels. For example, it improves the detection performance from $39.8$ to $42.6$ on the COCO dataset under $40\%$ box noise.

\begin{figure}[t]
    \centering
  \begin{subfigure}{0.4\textwidth}
    \includegraphics[width=0.9\linewidth]{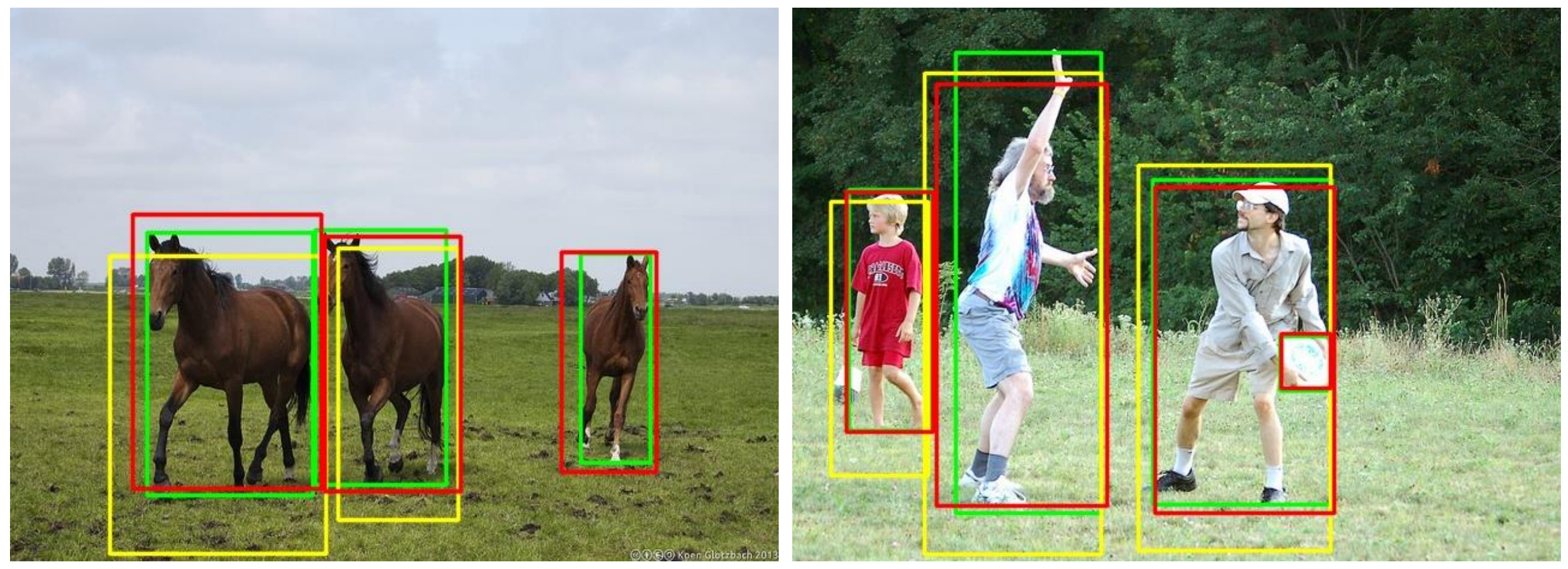}
	\caption{Qualitative results}
	\label{fig:det_vis}
  \end{subfigure}
  \quad
  \begin{subfigure}{0.4\textwidth}
    \centering
    \includegraphics[width=0.9\linewidth]{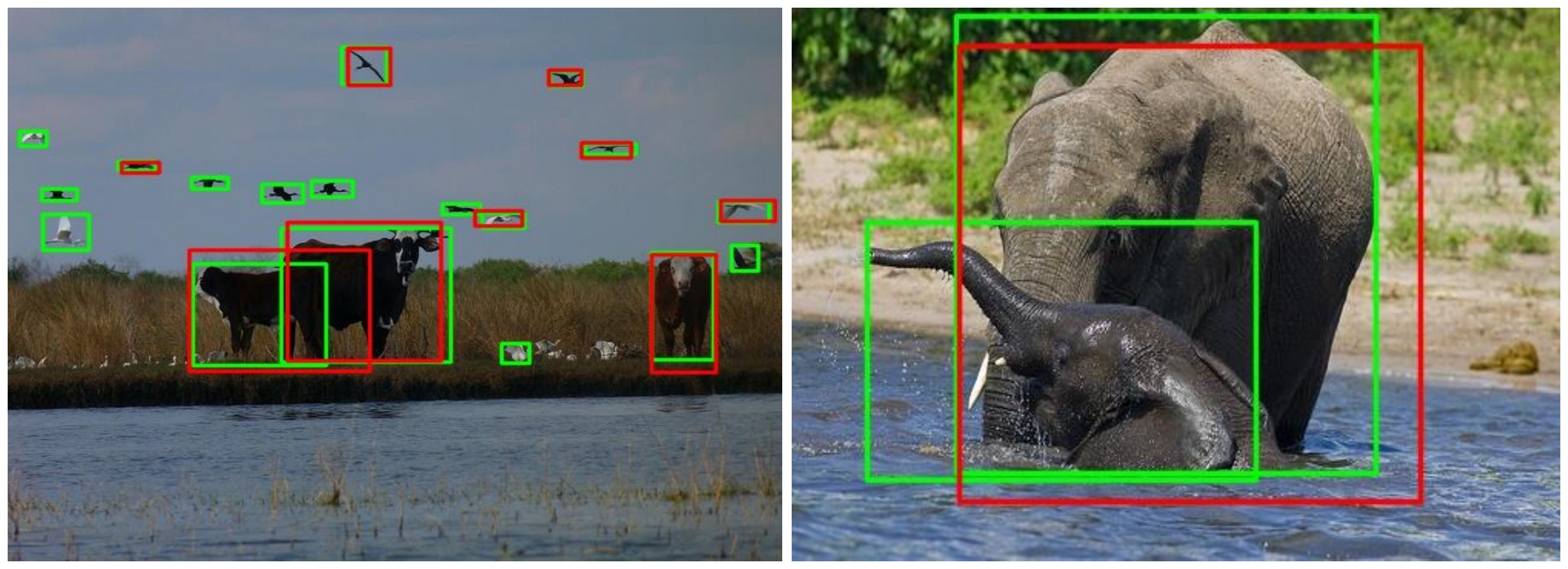}
    \caption{Failure cases} 
    \label{fig:failure}
  \end{subfigure}
\caption{(a) Qualitative results of OA-MIL FasterRCNN (red boxes) and vanilla FasterRCNN (yellow boxes) on the COCO dataset. The ground-truth boxes are in green. (b) Failure cases, \eg, missing detections on small/overlapped objects.}
\end{figure}

\medskip
\noindent \textbf{Qualitative Results.} 
Fig.~\ref{fig:det_vis} illustrates the qualitative results of the COCO dataset. The vanilla FasterRCNN tends to predict bounding boxes that cover object parts or include background areas. Instead, our method can predict more accurate bounding boxes. In addition, some failure cases are shown in Fig.~\ref{fig:failure}.
Our approach may suffer from overlapped objects or small objects.

\section{Conclusion}

In this work, we tackle learning robust object detectors with inaccurate bounding boxes.
By treating an object as a bag of instances, we present an Object-Aware Multiple Instance Learning method featured with object-aware instance selection and object-aware instance extension.
Our approach is general and can easily cooperate with modern object detectors. 
Extensive experiments on the synthetic noisy datasets 
and real noisy GWHD dataset demonstrate that OA-MIL can obtain promising results with inaccurate bounding box annotations.

For future work, we plan to incorporate the attributes of the objects to address the limitation of OA-MIL. 

\bigbreak
\noindent \textbf{Acknowledgement.}
This work was supported by the National Natural Science Foundation of China under Grant No. 61876211, No. U1913602, and No. 62106080.

\clearpage
%
%
\bibliographystyle{splncs04}
\bibliography{references}

\begin{thebibliography}{10}
\providecommand{\url}[1]{\texttt{#1}}
\providecommand{\urlprefix}{URL }
\providecommand{\doi}[1]{https://doi.org/#1}

\bibitem{bernhard2021}
Bernhard, M., Schubert, M.: Correcting imprecise object locations for training
  object detectors in remote sensing applications. Remote Sensing
  \textbf{13}(24) (2021)

\bibitem{bilen2015convex}
{Bilen}, H., {Pedersoli}, M., {Tuytelaars}, T.: Weakly supervised object
  detection with convex clustering. In: CVPR. pp. 1081--1089 (2015)

\bibitem{bilen2016deepmil}
{Bilen}, H., {Vedaldi}, A.: Weakly supervised deep detection networks. In:
  CVPR. pp. 2846--2854 (2016)

\bibitem{cascadercnn}
Cai, Z., Vasconcelos, N.: Cascade r-cnn: Delving into high quality object
  detection. CVPR pp. 6154--6162 (2018)

\bibitem{chadwick2019noisy}
{Chadwick}, S., {Newman}, P.: Training object detectors with noisy data. In:
  Proceedings of the IEEE Intelligent Vehicles Symposium (IV). pp. 1319--1325
  (2019)

\bibitem{chen2019mmdet}
Chen, K., Wang, J., Pang, J., Cao, Y., Xiong, Y., Li, X., Sun, S., Feng, W.,
  Liu, Z., Xu, J., Zhang, Z., Cheng, D., Zhu, C., Cheng, T., Zhao, Q., Li, B.,
  Lu, X., Zhu, R., Wu, Y., Dai, J., Wang, J., Shi, J., Ouyang, W., Loy, C.C.,
  Lin, D.: {MMDetection}: Open mmlab detection toolbox and benchmark. arXiv
  (2019)

\bibitem{cinbis2014mfold}
{Cinbis}, R.G., {Verbeek}, J., {Schmid}, C.: Multi-fold mil training for weakly
  supervised object localization. In: CVPR. pp. 2409--2416 (2014)

\bibitem{david2020wheat}
David, E., Madec, S., Sadeghi-Tehran, P., Aasen, H., Zheng, B., Liu, S.,
  Kirchgessner, N., Ishikawa, G., Nagasawa, K., Badhon, M.A., Pozniak, C.,
  de~Solan, B., Hund, A., Chapman, S.C., Baret, F., Stavness, I., Guo, W.:
  Global wheat head detection (gwhd) dataset: A large and diverse dataset of
  high-resolution rgb-labelled images to develop and benchmark wheat head
  detection methods. Plant Phenomics  \textbf{2020} (Aug 2020)

\bibitem{david2021wheat}
David, E., Serouart, M., Smith, D., Madec, S., Velumani, K., Liu, S., Wang, X.,
  Pinto, F., Shafiee, S., Tahir, I.S.A., Tsujimoto, H., Nasuda, S., Zheng, B.,
  Kirchgessner, N., Aasen, H., Hund, A., Sadhegi-Tehran, P., Nagasawa, K.,
  Ishikawa, G., Dandrifosse, S., Carlier, A., Dumont, B., Mercatoris, B.,
  Evers, B., Kuroki, K., Wang, H., Ishii, M., Badhon, M.A., Pozniak, C.,
  LeBauer, D.S., Lillemo, M., Poland, J., Chapman, S., de~Solan, B., Baret, F.,
  Stavness, I., Guo, W.: Global wheat head detection 2021: An improved dataset
  for benchmarking wheat head detection methods. Plant Phenomics  \textbf{2021}
  (Sep 2021)

\bibitem{deng2009imagenet}
{Deng}, J., {Dong}, W., {Socher}, R., {Li}, L., {Kai Li}, {Li Fei-Fei}:
  Imagenet: A large-scale hierarchical image database. In: CVPR. pp. 248--255
  (2009)

\bibitem{deselaers2010app}
Deselaers, T., Alexe, B., Ferrari, V.: Localizing objects while learning their
  appearance. In: ECCV. pp. 452--466 (2010)

\bibitem{diba2017casmil}
{Diba}, A., {Sharma}, V., {Pazandeh}, A., {Pirsiavash}, H., {Van Gool}, L.:
  Weakly supervised cascaded convolutional networks. In: CVPR. pp. 5131--5139
  (2017)

\bibitem{dietterich1997mil}
Dietterich, T., Lathrop, R., Lozano-Pérez, T.: Solving the multiple instance
  problem with axis-parallel rectangles. Artificial Intelligence  \textbf{89},
  31--71 (03 1997)

\bibitem{zisserman2010voc}
Everingham, M., Van~Gool, L., Williams, C.K.I., Winn, J., Zisserman, A.: The
  pascal visual object classes (voc) challenge. IJCV  \textbf{88}(2),  303--338
  (Jun 2010)

\bibitem{gao2019notercnn}
{Gao}, J., {Wang}, J., {Dai}, S., {Li}, L., {Nevatia}, R.: Note-rcnn: Noise
  tolerant ensemble rcnn for semi-supervised object detection. In: CVPR. pp.
  9507--9516 (2019)

\bibitem{aritra2017mae}
Ghosh, A., Kumar, H., Sastry, P.S.: Robust loss functions under label noise for
  deep neural networks. In: AAAI. p. 1919–1925 (2017)

\bibitem{ross2018detectron}
Girshick, R., Radosavovic, I., Gkioxari, G., Doll\'{a}r, P., He, K.: Detectron.
  \url{https://github.com/facebookresearch/detectron} (2018)

\bibitem{han2018coteach}
Han, B., Yao, Q., Yu, X., Niu, G., Xu, M., Hu, W., Tsang, I.W., Sugiyama, M.:
  Co-teaching: Robust training of deep neural networks with extremely noisy
  labels. In: NeurIPS. p. 8536–8546 (2018)

\bibitem{he2016resnet}
{He}, K., {Zhang}, X., {Ren}, S., {Sun}, J.: Deep residual learning for image
  recognition. In: CVPR. pp. 770--778 (2016)

\bibitem{he2019kl}
{He}, Y., {Zhu}, C., {Wang}, J., {Savvides}, M., {Zhang}, X.: Bounding box
  regression with uncertainty for accurate object detection. In: CVPR. pp.
  2883--2892 (2019)

\bibitem{jiang2018mentornet}
Jiang, L., Zhou, Z., Leung, T., Li, L., Fei-Fei, L.: Mentornet: Learning
  data-driven curriculum for very deep neural networks on corrupted labels. In:
  ICML. pp. 2309--2318 (2018)

\bibitem{vadim2016contextloc}
Kantorov, V., Oquab, M., Cho, M., Laptev, I.: Contextlocnet: Context-aware deep
  network models for weakly supervised localization. In: ECCV. pp. 350--365
  (2016)

\bibitem{alina2020openimagev4}
Kuznetsova, A., Rom, H., Alldrin, N., Uijlings, J., Krasin, I., Pont-Tuset, J.,
  Kamali, S., Popov, S., Malloci, M., Kolesnikov, A., Duerig, T., Ferrari, V.:
  The open images dataset v4. IJCV  \textbf{128}(7),  1956--1981 (Jul 2020)

\bibitem{li2016domain}
{Li}, D., {Huang}, J., {Li}, Y., {Wang}, S., {Yang}, M.: Weakly supervised
  object localization with progressive domain adaptation. In: CVPR. pp.
  3512--3520 (2016)

\bibitem{li2020noisy}
Li, J., Xiong, C., Socher, R., Hoi, S.C.H.: Towards noise-resistant object
  detection with noisy annotations. arXiv  \textbf{abs/2003.01285} (2020)

\bibitem{lin2017fpn}
{Lin}, T., {Dollár}, P., {Girshick}, R., {He}, K., {Hariharan}, B.,
  {Belongie}, S.: Feature pyramid networks for object detection. In: CVPR. pp.
  936--944 (2017)

\bibitem{lin2017retina}
Lin, T.Y., Goyal, P., Girshick, R., He, K., Dollár, P.: Focal loss for dense
  object detection. In: ICCV. pp. 2999--3007 (2017)

\bibitem{lin2014coco}
Lin, T.Y., Maire, M., Belongie, S., Hays, J., Perona, P., Ramanan, D.,
  Doll{\'a}r, P., Zitnick, C.L.: Microsoft coco: Common objects in context. In:
  ECCV. pp. 740--755 (2014)

\bibitem{liu2021iccvw}
Liu, C., Wang, K., Lu, H., Cao, Z.: Dynamic color transform for wheat head
  detection. In: ICCVW. pp. 1278--1283 (2021)

\bibitem{liu2022pp}
Liu, C., Wang, K., Lu, H., Cao, Z.: Dynamic color transform networks for wheat
  head detection. Plant Phenomics  \textbf{2022} (Feb 2022)

\bibitem{lu2022index}
Lu, H., Dai, Y., Shen, C., Xu, S.: Index networks. IEEE TPAMI  \textbf{44}(1),
  242--255 (2022)

\bibitem{ma2018d2l}
Ma, X., Wang, Y., Houle, M.E., Zhou, S., Erfani, S., Xia, S., Wijewickrema, S.,
  Bailey, J.: Dimensionality-driven learning with noisy labels. In: ICML. pp.
  3355--3364 (2018)

\bibitem{mao2020icip}
Mao, J., Yu, Q., Aizawa, K.: Noisy localization annotation refinement for
  object detection. In: ICIP. pp. 2006--2010 (2020)

\bibitem{ren2017fasterrcnn}
{Ren}, S., {He}, K., {Girshick}, R., {Sun}, J.: Faster r-cnn: Towards real-time
  object detection with region proposal networks. IEEE TPAMI  \textbf{39}(6),
  1137--1149 (2017)

\bibitem{siva2011drift}
{Siva}, P., {Tao Xiang}: Weakly supervised object detector learning with model
  drift detection. In: ICCV. pp. 343--350 (2011)

\bibitem{siva2012defence}
Siva, P., Russell, C., Xiang, T.: In defence of negative mining for annotating
  weakly labelled data. In: ECCV. p. 594–608 (2012)

\bibitem{song2019selfie}
Song, H., Kim, M., Lee, J.G.: {SELFIE}: Refurbishing unclean samples for robust
  deep learning. In: ICML. pp. 5907--5915 (2019)

\bibitem{song2014discovery}
Song, H.O., Lee, Y.J., Jegelka, S., Darrell, T.: Weakly-supervised discovery of
  visual pattern configurations. In: NeurIPS. p. 1637–1645 (2014)

\bibitem{tang2017online}
{Tang}, P., {Wang}, X., {Bai}, X., {Liu}, W.: Multiple instance detection
  network with online instance classifier refinement. In: CVPR. pp. 3059--3067
  (2017)

\bibitem{tang2016semitransfer}
{Tang}, Y., {Wang}, J., {Gao}, B., {Dellandréa}, E., {Gaizauskas}, R., {Chen},
  L.: Large scale semi-supervised object detection using visual and semantic
  knowledge transfer. In: CVPR. pp. 2119--2128 (2016)

\bibitem{tang2018semipami}
{Tang}, Y., {Wang}, J., {Wang}, X., {Gao}, B., {Dellandréa}, E., {Gaizauskas},
  R., {Chen}, L.: Visual and semantic knowledge transfer for large scale
  semi-supervised object detection. IEEE TPAMI  \textbf{40}(12),  3045--3058
  (2018)

\bibitem{uijlings2018semirevisit}
{Uijlings}, J.R.R., {Popov}, S., {Ferrari}, V.: Revisiting knowledge transfer
  for training object class detectors. In: CVPR. pp. 1101--1110 (2018)

\bibitem{uijlings2013ss}
Uijlings, J.R.R., van~de Sande, K.E.A., Gevers, T., Smeulders, A.W.M.:
  Selective search for object recognition. IJCV  \textbf{104}(2),  154--171
  (Sep 2013)

\bibitem{wan2018minent}
{Wan}, F., {Wei}, P., {Jiao}, J., {Han}, Z., {Ye}, Q.: Min-entropy latent model
  for weakly supervised object detection. In: CVPR. pp. 1297--1306 (2018)

\bibitem{wan2019cmil}
Wan, F., Liu, C., Ke, W., Ji, X., Jiao, J., Ye, Q.: {C-MIL:} continuation
  multiple instance learning for weakly supervised object detection. In: CVPR.
  pp. 2199--2208 (2019)

\bibitem{wei2018ts2c}
Wei, Y., Shen, Z., Cheng, B., Shi, H., Xiong, J., Feng, J., Huang, T.: Ts2c:
  Tight box mining with surrounding segmentation context for weakly supervised
  object detection. In: ECCV. pp. 454--470 (2018)

\bibitem{xu2021mrnet}
Xu, Y., Zhu, L., Yang, Y., Wu, F.: Training robust object detectors from noisy
  category labels and imprecise bounding boxes. IEEE TIP  \textbf{30},
  5782--5792 (2021)

\bibitem{zhang2019sdloc}
Zhang, X., Yang, Y., Feng, J.: Learning to localize objects with noisy labeled
  instances. In: AAAI. pp. 9219--9226 (2019)

\bibitem{zhang2019free}
Zhang, X., Wan, F., Liu, C., Ji, R., Ye, Q.: Freeanchor: Learning to match
  anchors for visual object detection. In: NeurIPS. pp. 147--155 (2019)

\bibitem{zhang2018gce}
Zhang, Z., Sabuncu, M.R.: Generalized cross entropy loss for training deep
  neural networks with noisy labels. In: NeurIPS. p. 8792–8802 (2018)

\end{thebibliography}
\end{document}